\documentclass{article}

\usepackage{PRIMEarxiv}

\usepackage[utf8]{inputenc} % allow utf-8 input
\usepackage[T1]{fontenc}    % use 8-bit T1 fonts
\usepackage{hyperref}       % hyperlinks
\usepackage{url}            % simple URL typesetting
\usepackage{booktabs}       % professional-quality tables

\makeatletter
\renewcommand\@biblabel[1]{#1.}
\makeatother

\usepackage{amsfonts} 
\usepackage{amstext}% blackboard math symbols
\usepackage{nicefrac}       % compact symbols for 1/2, etc.
\usepackage{microtype}      % microtypography
\usepackage{lipsum}
\usepackage{fancyhdr}       % header
\usepackage{graphicx}       % graphics
\graphicspath{{media/}}     % organize your images and other figures under media/ folder
\usepackage[labelfont=bf, figurename=Fig.]{caption}

\title{Individual Topology Structure of Eye Movement Trajectories
}

\author{
  Onuchin Arsenii*, Kachan Oleg \\
  Lomonosov Moscow State University, Moscow 119991, Russia*, \\
  Skolkovo Institute of Science and Technology, Moscow 143026, Russia.\\
  \texttt{onuchinaa@my.msu.ru*,} \texttt{oleg.kachan@skoltech.ru}\\
}

\begin{document}
\maketitle
\begin{abstract}
Traditionally, extracting patterns from eye movement data relies on statistics of different macro-events such as fixations and saccades. This requires an additional preprocessing step to separate the eye movement subtypes, often with a number of parameters on which the classification results depend. Besides that, definitions of such macro events are formulated in different ways by different researchers.

We propose an application of a new class of features to the quantitative analysis of personal eye movement trajectories structure. This new class of features based on algebraic topology allows extracting patterns from different modalities of gaze such as time series of coordinates and amplitudes, heatmaps, and point clouds in a unified way at all scales from micro to macro. We experimentally demonstrate the competitiveness of the new class of features with the traditional ones and their significant synergy while being used together for the person authentication task on the recently published eye movement trajectories dataset.

\keywords{topological data analysis, eye movement, eye tracking}
\end{abstract}
\section{Introduction}
Eye movement trajectories have been investigated and measured since the beginning of the twentieth century \cite{dodge1901angle}. From that moment, they have a long history of being used in many areas of psychology, cognitive science and applied research fields \cite{duchowski2017eye}. During such a long scientific history, the number of methods for quantitative characterization of eye movement trajectories did not change as dramatically as the number of scientific works with eye tracking methodology. Most of the quantitative measures present in the form of macro events statistics, where each eye movement trajectory decomposes into \textit{saccades and fixations}. Finding saccades in eye tracker data strongly associated with velocities calculation, usually using a very simple \textit{two-point central difference approach} \cite{inchingolo1985identification}. The simple sample-to-sample difference is the easiest way. However, this will significantly increase the noise of the signal \cite{bahill1982frequency}. So all of this mean, that the use of descriptive statistics based on macro events, assumes dependence on the chosen algorithm of velocity approximation and assumes extreme instability to noise in the signal in most cases. Therefore, the new properties, which can holistically describe the entire trajectory of eye movement, extract quantitative features from the entire trajectory. Topological features persist with any kind of random noise in the signal because of the holistic nature of these features. 

Eye movement trajectories are controlled by a vast and distributed neural system; therefore, they are affected by mental and neurological disorders such as attention deficit hyperactivity disorder (ADHD),  autism spectrum disorder (ASD) and Parkinson's disease \cite{Wang2015}. Therefore, the investigation of new features of eye movement trajectories has the potential to facilitate the diagnostic tasks of mental and neurological disorders. Other plausible areas to benefit from better features are ophthalmology \cite{Clark2019} and medical decision-making where eye movement trajectory data eliminates potential sources of error in image-based diagnostics \cite{Cai2018}. %and visual expertise used for expert skill assessment \cite{Cooper2009}.

% eye tracking data
Eye movement trajectory data are collected by eye trackers via emitting the infrared light into the viewer's eyes and capturing its reflection from the cornea, the transparent front part of the eye that covers the pupil. The point of gaze is then estimated from the relative positions of the pupil and the corneal reflection. In the most basic form, the eye tracker data is the time series of the coordinates of the eye movement trajectory $x$ and $y$ in pixels relative to the resolution of the screen. Additional information such as the pupil diameter and a vector of eye movement trajectory direction could be included depending on the tracker model and the experiment design.

In this work, we have investigated different topological features extracted from human eye movement trajectories and compared them with specific macro-event statistics during the solution of the classification problem.

\section{Materials and Methods}
\subsection{Statistics of fixations and saccades}

% fixation and saccades, clasifications -- thresholding, HMM, Kalman filter

Two macro-events that are prominent in most of the literature on eye movement trajectory are \textbf{fixations} and \textbf{saccades}. There are some periods or segments (with a fixed start point, end point, and duration) in which the eye movement trajectory trajectory is divided on occasion, supplemented by other events such as blinks or smooth pursuits. But even the definitions of such events are formulated in different ways by different researchers \cite{hessels2018eye}. Such a situation, when concepts as fundamental as saccades and fixations could differ according to scientific school, makes the investigation of alternative fundamental quantitative eye movement trajectory properties the part of current interest in the eye movement trajectory research field. 

\subsubsection{Fixations}
We are going to use definition of a fixation from \cite{leigh2015neurology} which sounds like: "Holds the image of a stationary object on the fovea by minimizing ocular drifts.". The fovea is the part of the eye that has a high concentration of cone cells that allow high-acuity vision. The fovea cover up a very small part of visual field, and the rest of the visual field (around the fovea) is low acuity vision, which means that the eye has to move itself to get an object into the fovea. There is no real maximum duration of a fixation, because we could control our eye movement trajectories, but they typically span in the interval $200$ - $400$ ms \cite{holmqvist2011eye}.

\subsubsection{Saccades}
Following the definitions from \cite{leigh2015neurology} lets find there definition of saccade: "brings images of objects of interest onto the fovea" . So saccades are very rapid and short eye movement trajectories with durations between $30$-$80$ ms, spanning amplitudes between $4$ deg - $20$ deg and velocity in interval from $30$ deg/s to $500$ deg/s \cite{holmqvist2011eye}. During fixations, objects of a scene are visually processed and information is encoded in memory, while during saccades visual input is suppressed. 

\subsubsection{Statistics}
In this section, we describe the quantitative properties of eye movement trajectories, extracted from track divisions to saccades and fixations. The statistics used in this work are also presented in Table \ref{tab:macro_features}. Let us define the track as $X = \{x_1, x_2, \dots, x_n\}$ and its partition into sets of fixations $X_f = \{f_1, f_2, \dots, f_k\}$ and saccades $X_s = \{s_1, s_2, \dots, s_k\}$, such that $\bigcup_{i \in I} f_i \cup \bigcup_{j \in J} s_j = X$ and $\bigcup_{i \in I} f_i \cap \bigcup_{j \in J} s_j = \emptyset$, where $I = |X_f|$ and $J = |X_s|$. Each individual fixation could be represented in the form $f_i = \{x_{1_i}, x_{2_i}, \dots, x_{m_i}\}_{i \in I}$ and similar for individual saccade $s_j = \{x_{1_j}, x_{2_j}, \dots, x_{m_j}\}_{j \in J}$.

If $X$ is the recording of the eye movement trajectory in some coordinate representation, with the corresponding splits in the sets of fixations $X_f$ and saccades $X_s$, then the obvious first-step features extracted from this split are \emph{numbers of fixations and saccades}, that is, $|X_s|$ and $|X_f|$. Also, for each fixation there is the fixed \emph{number of elements} in it, which is equal to $|f_i|$ and \emph{elements in all fixations} of the track which is equal to $|\bigcup_{i \in I} f_i|$. Similar procedure could be applied to saccades. 
Given a particular saccade $s_j = \{x_{1_j}, x_{2_j}, \dots, x_{m_j}\}_{j \in J}$, we can calculate the \emph{amplitude} of it and define it as the length of a vector between the first and last points that make up the saccade $s_j$.

\begin{equation}
    AMP_{s_j} = \| x_{1_j} - x_{m_j} \|.
\end{equation}

\emph{Integral elementary amplitude} is another characteristic which could be defined for each saccade $s_j$ and fixation $f_i$ as sum of set of elementary lengths 

\begin{equation}
    IAMP_{f_i} = \sum_{k\in |f_i|} \|x_k - x_{k+1}\|,~\quad\quad\quad\quad~IAMP_{s_j} = \sum_{k\in |s_j|} \|x_k - x_{k+1}\|.
\end{equation}

For the track itself, there is an integral elementary amplitude as the sum of IAMP for saccades and fixes. 

\begin{equation}
    IAMP = \sum_{i \in X} \| x_{i} - x_{i+1} \|
\end{equation}
 
The area of a convex hull $S(X)$ defined as the area of a convex polytope constructed on a finite set of data points $X \subset \mathbb{R}^2$. It is the unique convex polytope whose vertices belong to $X$ and surround all of $X$. 

Also we have used another quantitative parameter of a single track: \emph{peak amplitude}. It is a maximum amongst the elementary amplitudes of the saccade $\max \{\|x_i - x_{i+1}\|\}_{i\in |s_j|}$. 

Some of the macro-event statistics need additional standardization and vectorization before participating in the classification task. For example, set of saccade amplitudes per track couldn't be used directly in machine learning algorithms because of the unification need. For such a reason, we calculate a vector of min, max, mean, and standard deviation parameters for each of such sets. If the feature vector has a similar length for each experiment, then it can be used directly without any additional processing.  

\subsection{Topological data analysis (TDA)}

% TDA in 4 sentences
TDA \cite{Carlsson2009} introduces a new class of continuous vector-valued feature maps for a wide range of data domains such as \emph{point clouds}, \emph{time series} and \emph{scalar fields} summarizing the topology of the data in a multiscale way. 

\textbf{Point clouds} is a form of data representation, when elements of the data represents as unordered list of points in a Euclidean $n$-dimensional space $\mathbb{E}^n$. Following this definition we name a point cloud any finite subset of $\mathbb{E}^n$. Such a data form could be obtained from most of the natural experiments, even extracted from $2$-dimensional time series, after forgetting order on elements. The global topology of such point clouds could provide additional information about the structure of the data. 

\textbf{Time series} is an alternative form of data representation that has been used in the current work. Similarly, the $n$-dimensional time series is a point cloud with fixed linear order. Usually, time series have dimension $1$ or $2$. In this work, we work with $1$-dimensional time series, extracted from eye movement trajectory data. Such data capture the dependence of coordinates and amplitudes of gaze points and elementary eye movement trajectories on time. 

In addition, we have worked with \textbf{scalar fields}. It is a form of data representation, when to each element of point cloud have been attached some scalar value (usually from $\mathbb{R}$). For example, the value of the data point in the $2$-dimensional probability density distribution received from the kernel density estimation. Such scalar fields also named \textit{attention heatmaps} in an eye movement trajectories analysis context. 

The most common way to convert elements of the data cloud $\{x_i\} \subset \mathbb{E}^n$ to a single global topological object is to use these points as \emph{combinatorial graph} vertices. Edges of such a graph determined by some $\epsilon$ window of proximity -- when a distance between two points $\rho(x_i,x_j)$ less or equal to $\epsilon$, then there is an edge between $x_i, x_j$ vertices. This graph has a $2$-dimensional structure and cannot itself capture the high-dimensional properties of the original space, from which we distinguished the data points. For overcoming such problem there is a specific mathematical object, which could be created on any graph object --- \emph{clique complex} (specific method of simplicial complex creating). Each clique on $n$ verteces in graph will be interpreted as $n-1$-dimensional combinatorial simplex. TDA methods works directly with discrete constructions, but their topological properties could be generalized on topological realization of such combinatorial simplices --- \emph{topological simplices}. There are some different constructions of clique complexes, and the most useful and common in practice between them are the Vietoris-Rips (VR), Chech (C) and Alpha (A) complexes, see Figure \ref{fig:filtration} and Table \ref{tab:SC_vars}.

Converting point clouds into simplicial complex (VR, C or A) require choosing specific $\epsilon$ value. To avoid this problem, there is a way named \emph{filtration} described below.  

The central method of TDA, namely \emph{persistent homology} considers an ordered pair $(X, f)$ of a data points $X$ and a \emph{filter function} $f$ defined on the domain of interest $\mathcal{X}$. The filter function induces a \emph{filtration} of $X$ -- a sequence of subspaces of $X$

\begin{equation}
    \emptyset = X_0 \subseteq X_1 \subseteq X_2 \subseteq \dots \subseteq X_n = X,
\end{equation}

\begin{figure}
    \centering
    \includegraphics[width=\columnwidth]{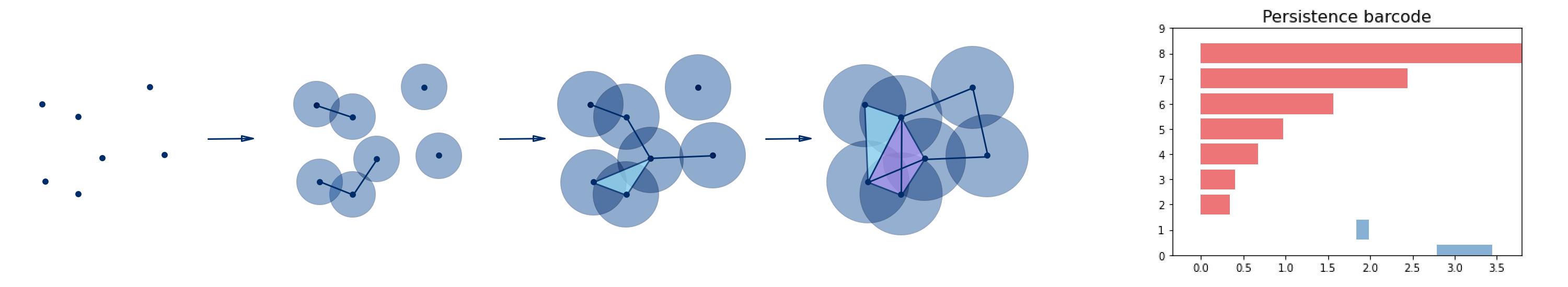}
    \caption{Vietoris-Rips clique complex generation process: from the separated point cloud elements to a high dimensional simplicial complex and corresponding persistence diagram on the right}
    \label{fig:filtration}
\end{figure}

which are often the sublevel sets $X_{\epsilon} = f^{-1}(-\infty, \epsilon]$ of this function for a threshold parameter $\epsilon$. For example, assume that $R = \{\text{VR}_i\}_1^N$ is a sequence of VR complexes associated with a point cloud (data) for an increasing sequence of parameter values $\{\epsilon_i\}_1^N$. 

\begin{equation}
    \text{VR}_1 \hookrightarrow \text{VR}_2  \hookrightarrow \cdots \hookrightarrow \text{VR}_N, 
\end{equation}

where $i$ --- inclusion maps between these complexes. And persistent homology provide a tool for examining homology not for a single complex  $\text{VR}_i$ but for a wholly sequence of homology groups in each dimension $*$ and for all $i < j$

\begin{equation}
    \iota: H^{*}(\text{VR}_i) \hookrightarrow H^{*}(\text{VR}_j).
\end{equation}

The ranks of homology vector spaces named Betti numbers $\beta_i = \text{rank}(H^{*}(X))$ play a role of the most common topological invariants in data analysis practice. 

\begin{table}
\begin{center}
\begin{tabular}{ |p{3cm}||p{10cm}|  }
 \hline
 \multicolumn{2}{|c|}{Cech - VietorisRips - Delaunay - Alpha complexes} \\
 \hline
 Cech & $ \text{Cech}_r(X) = \{\sigma \subseteq X \mid \bigcap_{x\in\sigma} B_r(x) \neq \emptyset\} $\\
 \hline
 VietorisRips   & $\text{VR}_r (X) = \{\sigma \subseteq X \mid \text{diam} (\sigma) \leq 2x\}$    \\
 \hline
 Delaunay &   $\text{Del}(X) = \{\sigma\subseteq X \mid \bigcap_{x\in\sigma} V_x \neq \emptyset \}$ \\
 & $V_x = \{y\in \mathbb{R}^d \mid ||y-x|| \leq ||y - z||, \forall z \in X\}$\\
 \hline
 Alpha & $\text{Alpha}_r (X) = \{\sigma \subseteq X \mid \bigcap_{x\in\sigma}(B_r(x) \cap V_x) \neq \emptyset\}$ \\
 \hline
\end{tabular}
\caption{Formal Definitions of Different Simplical Complex Constructions}
\label{tab:SC_vars}

\end{center}
\end{table}

While the threshold parameter is changed from $-\infty$ to $+\infty$ persistent homology allows tracking the appearance and disappearance of topological features of various dimensions, i.e. $0$-dimensional topological features are connected components, $1$-dimensional features are holes, $2$-dimensional are voids, etc. A topological feature is considered \emph{persistent} if it exists for a long interval of the threshold parameter. The \emph{persistence} of $i$-th topological feature is the difference between its \emph{death} (disappearance) $d_i$ and \emph{birth} (appearance) $b_i$ times -- the values of the filter function $f$ at $\epsilon$.

% persistent diagram
As a result of a filtration for a chosen maximum dimension $K$ one obtains a collection $\{D^k_f(X)\}_{k \in \{0, \dots, K\}}$ of compact descriptors of the data called a \emph{$k$-th dimensional persistence diagram} (PD) $D_f^k(X) = \{(b_i, d_i)\}_{i \in I}^k$ which is the multiset of birth-death intervals of topological features of a dimension $k$. Equivalently, a persistence diagram is a multiset of points on the extended Euclidean plane $\mathbb{R}^2 \cup \{+\infty \}$ in the birth-death coordinates. Furthermore, we exclude the dependence of a persistence diagram $D^k$ on $X$ and $f$ to be inferred from the context. A number of software packages -- \texttt{Dionysus}, \texttt{GUDHI}, \texttt{Ripser} \cite{Maria2014, Bauer2021} exists to effectively compute the PDs of data in different domains.

% data modalities, Vietoris-Rips filtration, sublevel set filtration

As mentioned above, we consider three different data representations: point clouds, time series, and scalar fields. Persistent homology was calculated for each of these representations. We endow this modalities with Vietoris-Rips (VR) and $2$D and $1$D sublevel set filtrations respectively. The former two allow to quantify the complexity of the eye movement trajectory distribution support, while the latter one is a more principled approach to summarize the ad-hoc features such as number of saccades, their duration and amplitude. For rigorous treatment, we refer the reader to \cite{Carlsson2009,Edelsbrunner2010} due to the lack of space.

%\input{figures/figure2} % sublevel set filtration

% vector representations of persistent diagrams
As the persistence diagram is a multi-set of intervals and is not an element of a vector space, it should be transformed into a vector to be used in machine learning algorithms. We use the piecewise-linear functions of the persistence diagrams, namely the \emph{betti curve}, \emph{persistence curve} and \emph{persistence entropy curve} evaluated on a fixed grid with $n$ knots to obtain a $n$-dimensional vector.

\emph{Betti curve} summarize all homology ranks for a single filtration process and present them in vector form. 

\begin{equation}
    \beta_i(\epsilon) = \text{rank}(H^{i}(VR_{\epsilon}))
\end{equation}

\emph{Persistence curve} \cite{Chung2019} summarizes the total persistence of topological features at the time $\epsilon$ of a filtration and is defined.

\begin{equation}
    P_{D^k}(\epsilon) = \sum_{i=1}^{|D^k|} (d_i - b_i),~~~\mathrm{s.t.}~b_i < \epsilon < d_i.
\end{equation}

Consider a set of piecewise linear functions $\{\Lambda_p^k(\epsilon)\}_{p \in D^k}$ for all birth-to-death pairs $p = (b, d) \in D^k$ as $\Lambda_p(\epsilon) = \max(0, \min(\epsilon - b, d - \epsilon))$, then \emph{persistence landscape} \cite{Bubenik2017} $L_{D^k}$ is defined as 

\begin{equation}
    L^n_{D^k}(\epsilon) = \textrm{nmax}_p \Lambda_p(\epsilon).
\end{equation}

\subsection{Dataset}

% dataset
In our study we used a subset of GazeBase dataset \cite{Griffith2021} acquired in the Texas State University, TX, USA. It consists of $1260$ eye movement trajectory recordings from $14$ participants. Each person participated in $9$ rounds with $2$ sessions in each performing $5$ different visual stimuli tasks: fixation (\texttt{FIX}), horizontal saccade (\texttt{HS}), random saccade (\texttt{RS}), text reading (\texttt{TEXT}) and gaze-driven gaming (\texttt{GAME}).

% ethics statement
%All participants provided informed consent under a protocol approved by the Institutional Research Board at Texas State University prior to each round of recording. This IRB review process was guided by the ethical principles specified in the Belmont Report and by regulations of the U.S. Department of Health and Human Services. As part of the consent process, participants acknowledged that the resulting data may be disseminated in a de-identified form.

% acquisition parameters
Monocular eye movement trajectories data were acquired using an EyeLink $1000$ eye tracker with $1$ ms temporal resolution. Stimuli were presented to participants on a $1680 \times 1050$ pixel ($474 \times 297$) ViewSonic monitor. Participants completed two sessions of recording for each round of collection. The gaze position and corresponding stimuli were innately expressed in terms of pixel display coordinates. These values were converted to degrees of visual angle (DVA) according to the geometry of the recording setup. For further details reader is advised to refer to the original article \cite{Griffith2021}.

% tasks overview, each person participated 
Each person participated in five different tasks, which we outline below. For the first three tasks, a target, a white point, was displayed on a black background, starting at the center of the screen. For the fixation (\texttt{FIX}) task, a target remained in the center of the screen for a duration of $15$ seconds. For the horizontal saccade (\texttt{HS}) task, a target was repeatedly switched between two horizontally located positions at $\pm 15^{\circ}$ of the visual angle at intervals of $1$ seconds for a duration of $101$ seconds. For the random saccade (\texttt{RS}) task, a target changed positions at random within a rectangular region of $\pm 15^{\circ}$ horizontally and $\pm 9^{\circ}$ vertically with a minimum amplitude of $2^{\circ}$ at one-second intervals for a duration of $101$ seconds. For the text reading (\texttt{TEXT}) task an excerpt of a poem in English language specific for each session was displayed for $60$ seconds with white font on a black background. For the game (\texttt{GAME}) task, red and blue balls were displayed moving along a random linear trajectory at constant speed, bouncing off the edges of the screen. The participant's goal was to pop all of the red balls by fixating a gaze on them, while blue balls were unpoppable. This task lasted until a participant popped all of the red balls.

% data format
For each track, the data consist of the $x$ and $y$ coordinates of the gaze, the timestamp, and a label relating an elementary event to a fixation, saccade, or blink provided by the built-in EyeLink parser.

\subsection{Experiment}

The original raw data \cite{Griffith2021} consists of a $2$ D time series of $(x_t,y_t)$ coordinates of gaze at a specific moment of time $t$. Furthermore, we had pre-processed information on $X_f$ and $X_s$ in each gaze. 

\begin{figure}
    \centering
    \includegraphics[width=\columnwidth]{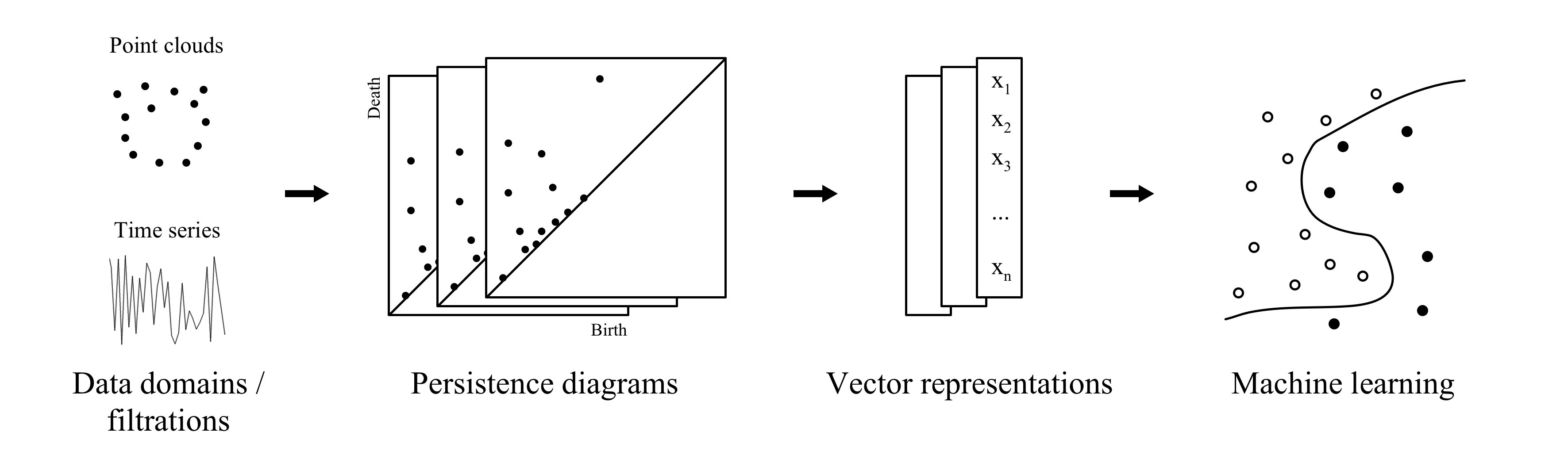}
    \caption{General pipeline description. First step is a specific data representation extraction. Second step is a topological features extraction. Third step consists in a vectorization of extracted topological features. And a final step is a machine learning algorithm realization}
    \label{fig:bio_int}
\end{figure}

For quantitative estimation of the new proposed class of topological features, we considered the $14$-class classification problem to discriminate between the eye movement trajectories obtained for the $5$ different tasks, which constitution was discussed in the previous Data section.

%% Here needed explanation of new filtration methods which will be used
On the noise removal stage, the high amplitude error points were removed from eye movement trajectory. Linear interpolation was used to remove and fill in empty spaces and gaps in the tracks. Furthermore, the raw data was used to construct visual scanpaths represented by a matrix $\mathbf{X}_i \in \mathbb{R}^{n_i \times 2}$, where $n_i$ is the number of gaze points within the $i$-th scanpath.

\begin{figure}
    \centering
    \includegraphics[width=\columnwidth]{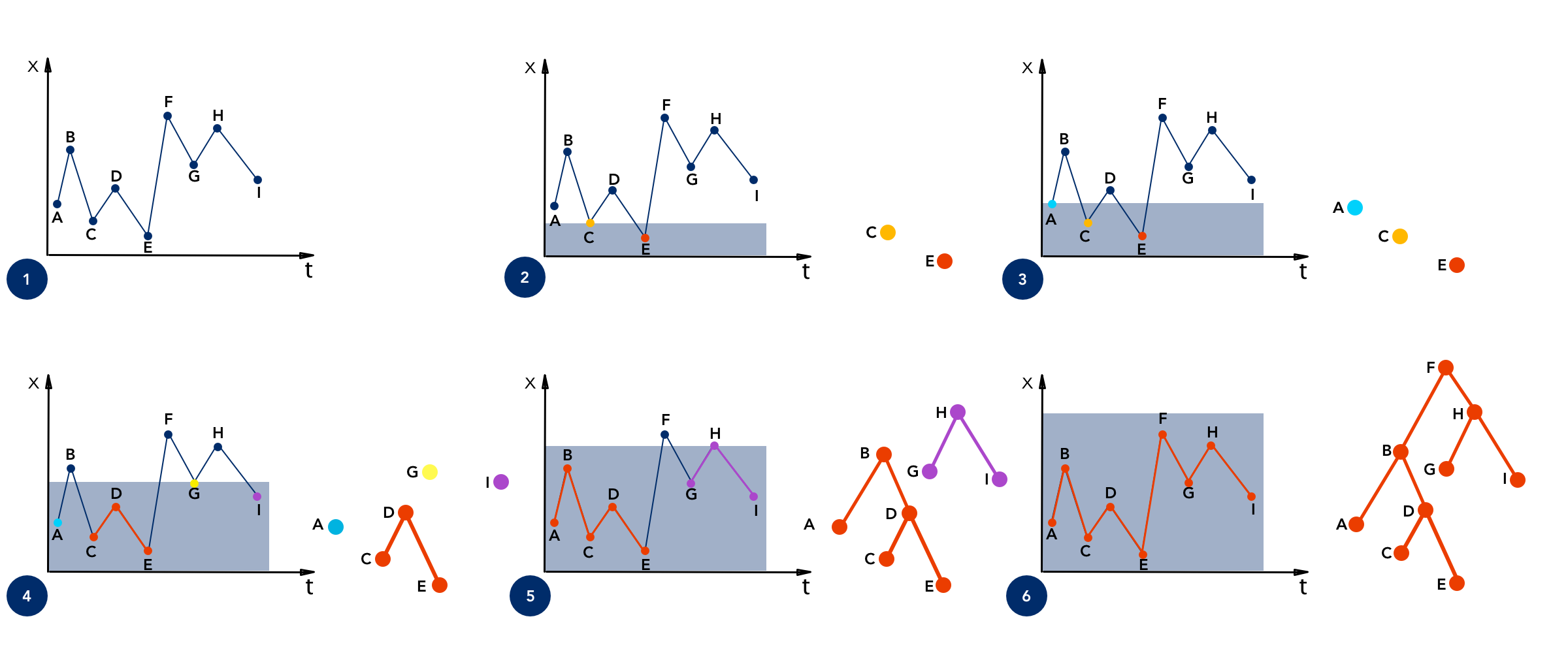}
    \caption{One dimensional time series filtration process, which named \textit{lower star filtration} with the corresponding simplicial complexes}
    \label{fig:filtration}
\end{figure}

For each $\mathbf{X}_i$ different feature vectors were computed. For the baseline, we computed different statistics from macro-events: fixations, saccades, and track itself. And vectorize some of these statistics, which do not have unified form for each experiment, by computing the min, max, mean, and standard deviation parameters. 

As topological features, we consider the analysis of $1$D time series per each coordinate of the scanpath --- $x$ and $y$. 

The connectivity of such $1$D time series of functional dependence between scanpath coordinates on the time index of the eye movement trajectory was computed using the lower star filtration, resulting in $D^0$ persistent diagram, see Figure \ref{fig:filtration}. Intuitively, this kind of topological feature summarized the number of saccades, their duration, and amplitude along a specified axis during data acquisition. Note that we computed both the lower and upper filtrations of the time series by flipping its sign.

The diagrams were vectorized betti and entropy curves \cite{atienza2020stability} of dimension $100$, implemented by \texttt{Gudhi} and fed to the random forest  classification algorithm.

The data was split into train and test subsets with the $80/20$ ratio. The mean and standard deviation of the accuracy metric averaged over $75$ runs are reported.

\begin{table}
\begin{center}
    \caption{Horisontal saccade (\texttt{HS}), Random saccade (\texttt{RS}),  Text reading (\texttt{TEXT}), Game (\texttt{GAME}), Fixation (\texttt{FIX}),  Accuracy of individual features of standard statistics of fixations and saccades along with multi-modal combined features: topological feature plus standard statistics and their total combination.}
    \begin{tabular}{|p{1cm}|l||lllll|}
    \hline
        & \textbf{Standard}   & \multicolumn{5}{c|}{\textbf{Time series}}  \\
    \textbf{Task}                                & \textbf{Statistics}    & \textbf{X}         & \textbf{Y}         & \textbf{Amp }      &\textbf{ Landscape} & \textbf{All }   \\ \hline
    \texttt{FIX} & 0.4107        & 0.4156    & 0.3496    & 0.4304 & 0.4201    & \textbf{0.5087}   \\
    \hline
    \texttt{HS}  & 0.7333        & 0.5696    & 0.4512    & 0.5274 & 0.5647   & \textbf{0.8034}    \\
    \hline
    \texttt{RS} & \textbf{0.6489}        & 0.2305    & 0.2357    & 0.4842 & 0.4405   & 0.5663       \\
    \hline
    \texttt{TEXT}   & \textbf{0.7657}        & 0.3458    & 0.2616    & 0.4670  & 0.4193  & 0.6898  \\
    \hline
    \texttt{GAME}   & 0.4862       & 0.2016    & 0.2334    & 0.3952  & 0.3793  & \textbf{0.5277}     \\ 
    \hline
        &    & \multicolumn{5}{c|}{\textbf{Multi-modal}}  \\
                                   &     &          &          &        &  &     \\ \hline
    \texttt{FIX} & 0.4107 & 0.4456    & 0.4410    & 0.5172  & 0.4823   & \textbf{0.5586}   \\
    \hline
    \texttt{HS}  & 0.7333 & 0.7504    & 0.7227    & 0.7611 & 0.7023   & \textbf{0.8263}   \\
    \hline
    \texttt{RS} & 0.6489 & 0.6888    & 0.6853    & \textbf{0.6932}  & 0.6694  & 0.6762     \\
    \hline
    \texttt{TEXT}  & 0.7657  & 0.7522    & 0.7555    &  0.7729  & 0.7843  & \textbf{0.8015}   \\
    \hline
    \texttt{GAME}  & 0.4862 & 0.4377    & 0.4643    & \textbf{0.5809}  & 0.4538  & 0.5576      \\ 
    \hline
    \end{tabular}
\label{tab:res_comb}
\end{center}
\end{table}

\section{Results}

It was clearly shown that standard feature statistics have an implicit synergy with topological features and their combinations, what result in growth of classifier accuracy in Table \ref{tab:res_comb}. The solution of the classification problem by methods of only topological features has also shown superiority over only standard macro-event features in FIX, HS and GAME tasks. Especially good results have been shown in the HS, so it can be assumed that topology is somehow related to the individual nature of the participants (accuracy $=0.8034$ for combination of only topological features). Also TEXT reading task has shown significant improvement in combination of all features together (accuracy $=0.8015$). Thereby, we can suggest that the information summarized by concatenate feature representations is correlated to some extent, can support a more general understanding of the nature of eye movement trajectories, and can be used in classification problems of any kind. 

As a conclusion it was shown that the topological characteristics of eye movement can be effectively used in classification problems and have synergistic properties with wide range of macro event features. Because of the independence that topological features have from velocity calculation and any kind of local noise in the data, such a pipeline could be effectively used in addition to the standard pipeline or separately from it.

\bibliographystyle{unsrt}  
\bibliography{references}

\appendix

\section{Additional definitions and notation}

\begin{table}[h]
\begin{center}
\caption{Macro-event features description}
\begin{tabular}{ |p{8cm}||p{6cm}|  }
 \hline
 \multicolumn{2}{|c|}{\textbf{Fixations}} \\
 \hline
 Number of fixations in track & $|X_f|$\\
 \hline
 Lengths of fixations (number of elementary events)   & $\{|f_i|\}_{i\in I}$    \\
 \hline
 Integral elementary amplitude of each fixation &   $\sum_{i\in |f_j|} \|x_i - x_{i+1}\|$\\
 \hline
 \multicolumn{2}{|c|}{\textbf{Saccades}} \\
 \hline
 Number of saccades in track & $|X_s|$\\
 \hline
 Length of saccades (number of elementary events)   & $\{|s_j|\}_{j\in J}$   \\
 \hline
 Amplitude &   $\|x_{1_j} - x_{m_j} \|$ for each $s_j$\\
 \hline
 Integral elementary amplitude of each saccade &   $\sum_{i\in |s_j|} \|x_i - x_{i+1}\|$\\
 \hline
 Peak amplitude of each saccade &   $\max \{\|x_i - x_{i+1}\|\}_{i\in |s_j|}$\\
 \hline
 \multicolumn{2}{|c|}{\textbf{Track}} \\
 \hline
 Integral elementary amplitude & $\sum_i^{|X|} \|x_i - x_{i+1}\|$\\
 \hline
 Area of convex hull  & $S(X)$   \\
 \hline
\end{tabular}
\label{tab:macro_features}
\end{center}
\end{table}

\end{document}